\documentclass{article}

\usepackage{amssymb}
\usepackage{amsthm}
\usepackage{amsmath}
\usepackage{microtype}
\usepackage{graphicx}
\usepackage{subfigure}
\usepackage{booktabs} % for professional tables

\usepackage{hyperref}

\usepackage[accepted]{icml2021}

\icmltitlerunning{Predictive Representation Learning for Language Modeling}

\usepackage{amssymb}
\usepackage{amsmath}
\usepackage{multirow}
\usepackage{graphicx}
\usepackage{subfigure}
\usepackage{booktabs} % for professional tables
\usepackage{paralist} % for compact items
\usepackage{bbm}      % for indicator function
\usepackage{color}
\usepackage{paralist} % for more compact lists

\newcommand{\Actions}{\mathcal{A}}
\newcommand{\States}{\mathcal{S}}

\begin{document}

\twocolumn[
\icmltitle{Predictive Representation Learning for Language Modeling}




\begin{icmlauthorlist}
\icmlauthor{Qingfeng Lan}{to}
\icmlauthor{Luke Kumar}{goo}
\icmlauthor{Martha White}{to}
\icmlauthor{Alona Fyshe}{to}
\end{icmlauthorlist}

\icmlaffiliation{to}{Department of Computing Science, University of Alberta, Edmonton, Alberta, Canada}
\icmlaffiliation{goo}{Alberta Machine Intelligence Institute, Edmonton, Alberta, Canada}
\icmlcorrespondingauthor{Qingfeng Lan}{qlan3@ualberta.ca}

% The \author macro works with any number of authors. There are two commands
% used to separate the names and addresses of multiple authors: \And and \AND.
%
% Using \And between authors leaves it to LaTeX to determine where to break the
% lines. Using \AND forces a line break at that point. So, if LaTeX puts 3 of 4
% authors names on the first line, and the last on the second line, try using
% \AND instead of \And before the third author name.

\icmlkeywords{language modeling, predictive representation, reinforcement learning}

\vskip 0.3in
]



\printAffiliationsAndNotice{}  % leave blank if no need to mention equal contribution

\begin{abstract}

To effectively perform the task of next-word prediction, long short-term memory networks (LSTMs) must keep track of many types of information. Some information is directly related to the next word's identity, but some is more secondary (e.g. discourse-level features or features of downstream words). Correlates of secondary information appear in LSTM representations even though they are not part of an \emph{explicitly} supervised prediction task. In contrast, in reinforcement learning (RL), techniques that explicitly supervise representations to predict secondary information have been shown to be beneficial.
Inspired by that success, we propose Predictive Representation Learning (PRL), which explicitly constrains LSTMs to encode specific predictions, like those that might need to be learned implicitly. We show that PRL 1) significantly improves two strong language modeling methods, 2) converges more quickly, and 3) performs better when data is limited. Our work shows that explicitly encoding a simple predictive task facilitates the search for a more effective language model.
\end{abstract}

\section{Introduction}

Language generation is a complex task. When generating the next word in a sentence, there are multiple things to keep track of (style of writing, topic of the document, current point in a sentence's structure, etc.), and yet it remains a highly unconstrained problem. There are many equally suitable ways to say the same thing. A common solution to the language generation problem is to train a Recurrent Neural Network (RNN) to predict the next word in a sequence, conditioned on the previous words.

An effective RNN must retain relevant past information, and also provide useful features to predict what is likely to happen beyond the next word. In multiple instances, the hidden states of RNNs have been shown to encode predictions about upcoming words, including plurality, grammatical number information, and subject-verb dependency \citep{lakretz2019emergence}. But RNNs are not given any guidance on what to remember, nor what predictions to encode (beyond next word identity). This makes the learning problem much more difficult, and difficult problems tend to require more data.

Predicting future events is a key component in reinforcement learning (RL).    
Value functions in RL encode predictions about future rewards, as well as predictions secondary to the main reward~\citep{sutton2011horde}. Models with representations that are forced to encode secondary predictions generalize better~\citep{littman2002predictive,schaul2013better,matthew2018gvfn}, a concept known as the \emph{predictive representation hypothesis} \citep{rafols2005using}. %This technique has been successful in robotics and time series prediction \citep{matthew2018gvfn,modayil2012multi,white2015developing}. 

Inspired by this, we propose a new approach to language modeling: Predictive Representation Learning (PRL). PRL is a sequence-to-sequence framework designed for auxiliary task learning. PRL generates predictive representations while learning to solve a simple sequence labeling task (e.g. part-of-speech tagging or named entity recognition). These PRL representations are then incorporated into the hidden representations of a more complex language model, allowing the language model to leverage those predictions when selecting the next word. % I don't love that this is predictions followed by predicting but I can't think of something better right now.
Our work clearly demonstrates that optimizing for a simple task and incorporating those predictions into the hidden states, help the model to find better solutions to the much more complex problem of language modeling. 

To summarize, our main contributions are:
% \begin{itemize}
\begin{compactenum}
%\item an example of how to solve sequence labeling tasks with an RL algorithm;
\item PRL, an auxiliary task framework that learns predictive hidden representations for language modeling; 
\item evidence that PRL improves language modeling performance perplexity by up to 5\% ;
\item evidence that PRL improves both the convergence rate and sample efficiency of language models.
\end{compactenum}
% \end{itemize}

\section{Related Work}

Our work is closely related to predictive representations and multi-task learning, as we outline in this section.

\subsection{Predictive Representations}

The predictive representations hypothesis, first proposed in the area of RL, posits that representations that predict future observations are better for generalization~\cite{rafols2005using}. This hypothesis was inspired by earlier work, 
which developed predictive state representations (PSRs) \citep{littman2002predictive}. A PSR is a vector of predictions about the probabilities of several future action-observation sequences. Later, the state prediction took the form of value functions. \citet{sutton2005temporal} extended Temporal-Difference (TD) learning methods and proposed TD networks as an effective alternative learning algorithm for PSRs. 

Most recently TD networks were generalized, to highlight a connection to RNNs where the hidden state is directly constrained to be value function predictions \citep{matthew2018gvfn}.
They showed that constraining the hidden state to be predictions reduced sensitivity to the truncation level in back propagation through time, and often improved convergence rates. Our work is inspired by this insight, where instead of RL, we investigate the specification and utility of predictive representations for language modeling.

% IS it worthwhile to talk about tests? Also, i would say value functions are tests
%%These sequences are also known as \emph{tests}. 
%%Among all tests, \emph{core tests} are those tests that are sufficient to make predictions about the future. 
%For each test, an agent records the success probability as a feature in the state representation of the agent.
%
%Value functions can also be viewed as special forms of tests. \citet{sutton2005temporal} extended Temporal-Difference (TD) learning methods and proposed TD networks as an effective alternative learning algorithm for PSRs. 
%% TD networks include two kinds of networks: question networks and answer networks. A question network suggests a question about possible future observations, analogous to a test in PSR. An answer network learns to answer the question, analogous to the function used to compute the success probability of the test.
%The predictions of TD networks are incorporated into state representations. \citet{sutton2005temporal} showed that TD networks can learn state representations and help to find an exact solution for a non-Markov problem.

Predictive representations have also proved their value in the area of NLP. \citet{trinh2018learning} applied unsupervised auxiliary losses to force RNNs to reconstruct past sequences or predict future sequences. These losses regularize representations in RNNs, which significantly improve the optimization and generalization of RNNs.

Predictive representations are also beneficial for large models that exploit large amounts of data.
\citet{kuncoro2020syntactic} proposed structure-distilled BERT models by injecting explicit syntactic inductive biases into BERT models. The syntactic biases are expressed as syntactically informative predictions, in the form of probability distributions. The new models outperformed the baseline on six diverse tasks.

\subsection{Multi-task Learning}

There are many multi-task learning frameworks for NLP tasks. In these frameworks, parameters for each task can be shared among tasks, improving the optimization of those shared parameters. \citet{bingel2017identifying} found that multi-task learning can help neural networks training get out of local minima. Multi-task learning also improves model generalization, accelerates the training process, and allows for knowledge sharing across domains \cite{caruana1997multitask, ruder2017overview}.  

Specifically, \citet{luong2015multi} developed a multi-task sequence-to-sequence model and found that improves translation quality between English and German when trained for syntactic parsing and image caption.
%\citet{sogaard2016deep} presented a multi-task learning RNN architecture for sequence tagging. When comparing low-level task supervision (e.g. POS tagging) at both the innermost and outmost layers of the RNN, \citet{sogaard2016deep} found that supervision at the innermost layer gives the greatest increase in performance. 
Furthermore, \citet{hashimoto2017joint} proposed a joint many-task model that reflects linguistic hierarchies and achieves state-of-the-art or competitive results on many tasks.
Similarly, \citet{sanh2019hierarchical} introduced a hierarchical model trained by supervising low-level tasks at lower layers and more complex tasks at higher layers. More diverse semantic information is encoded in the learned representations.
\citet{subramanian2018learning} applied multi-task learning to learn general purpose distributed sentence representations. They demonstrated that these representations greatly speed up low-resource learning and they are competitive to or better than other sentence representations produced by previous methods.

In contrast to these approaches, our work focuses on incorporating predictions into states for language modeling. This differs from indirectly using auxiliary tasks.

\section{Predictive Representations for Language Modeling}

We begin with an intuitive example showing how predictions can improve language modeling (LM) by significantly reducing the hypothesis space. Then, we present two different forms of predictive representations (P and Q predictions) for LM.

\subsection{Predictions Reduce the Hypothesis Space}

One of the challenges in LM is a large vocabulary set over which the model must produce predictions.  For example, language models trained in English often have a vocabulary of tens of thousands of words.  Choosing one word from this large set contributes to the difficulty of the problem.

PRL is inspired by the idea that LM would be easier if we had some additional information about the next word. LM is strongly related to part-of-speech (POS) tagging~\cite{heeman1999pos} and named entity recognition (NER)~\cite{sachan2018effective}. With access to the predicted POS/NER tag of the next word, a language model may be able to reduce the number of probable next words, thus improving accuracy. 

\begin{figure}[tb]
\vskip 0.2in
\begin{center}
  \centerline{\includegraphics[width=0.8\columnwidth]{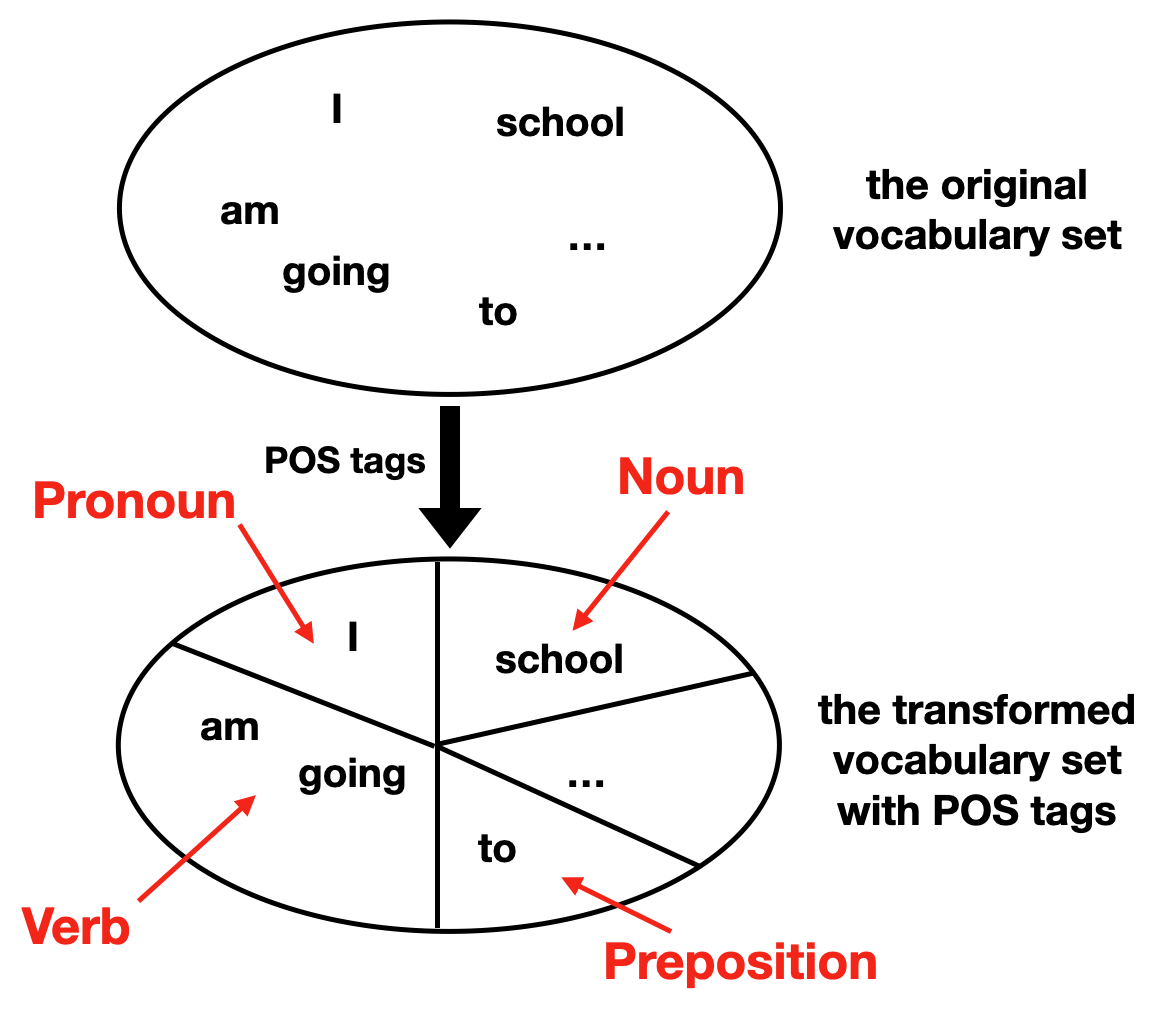}}
  \caption{The predictions of POS tags help reduce the hypothesis space greatly. }
  \label{fig:motivation}
\end{center}
\vskip -0.2in
\end{figure}

For example, if we knew the next word was a noun, that would reduce our prediction space to nouns only.  POS tags partition the prediction space and allow us to focus on only a subset of the total vocabulary when making our next-word prediction (see Figure~\ref{fig:motivation}).  Predicting the POS tag of the next word is an easier task than predicting the next word, and so the predictions are fairly trustworthy. Thus, the secondary POS prediction task can provide an additional explicit signal to a language model. 

PRL forces the representations of the language model to be predictions for these supervised sequence labeling tasks, which can be leveraged by the language model. Many such sequence labeling tasks could be considered; here we explore POS tagging and NER.%hypothesis space actually has a special meaning in ML and I don't think it's a good idea to use it here with different intent.

%Thus part-of-speech (POS) tags as a piece of useful information, but any relevant information is applicable in general. Assume that there is a large vocabulary containing all English words. Given the POS tags of these words, we can divide this large vocabulary set into several smaller subsets. Each subset contains all and only words with a certain POS tag. Predicting the next word is hard while predicting the POS tag of the next word is much easier since the size of the POS tag set is much smaller than the size of the vocabulary set. Once knowing the predicted POS tag of the next word, we choose a word only from the subset with the predicted POS tag. In other words, the hypothesis space is reduced greatly with the help of predictions.

%
% In particular, we study two forms of predictive representations and manage to combine them with language models.

\subsection{Two Forms of Predictive Representations}

How can we communicate the predictions from a sequence labeling model to a language model?  One possibility is to pass the raw probabilities from the classifier to the language model. For a classification task such as POS tagging, a supervised model usually generates a probability distribution over all POS tags. We call these probabilities \textbf{P predictions}.

\begin{figure}[tb]
\vskip 0.2in
\begin{center}
  \includegraphics[width=\columnwidth]{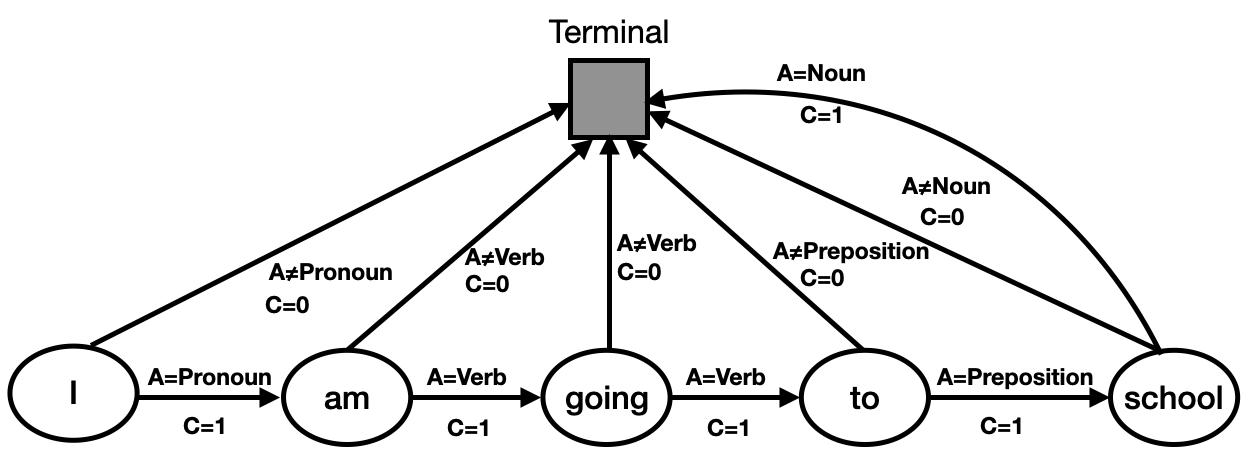}
  \caption{An MDP representation of the sentence example for POS tagging. Only when the correct label (A) is selected does the MDP move to the next state, receiving a reward (C) of 1. All incorrect labels result in a transition to the terminal state with zero reward.}
  \label{fig:mdp}
\end{center}
\vskip -0.2in
\end{figure}

The second form of predictive representations we consider is the output of action-value functions from RL. We first transform a sequence labeling task (POS tagging) into an RL problem, which has been studied previously~\cite{keneshloo2019deep}.
``I am going to school." is the example sentence we want to label. The correct POS labels are ``I (Pronoun) am (Verb) going (Verb) to (Preposition) school (Noun)."
We set the state space $\States$ to be the vocabulary set, i.e. $\States=\{I, am, going, to, school, \dots \}$. The action space $\Actions$ is the set of all possible POS labels, such as $\Actions=\{Pronoun, Verb, Preposition, Noun, \dots \}$.
The example sentence together with the correct labels can be represented as a finite Markov Decision Process (MDP) as shown in Figure \ref{fig:mdp}. 
When a correct POS label for the current word is chosen, it leads to the next state with a reward of 1; otherwise, it goes to the terminal state with a reward of 0.
For example, in state ``am" if action ``Verb" is taken, the agent goes to the next state (``going''); otherwise it transitions to the terminal state. 

% Specifically, the state space $\States$ is the vocabulary set, and the action space $\Actions$ is defined as the set of all possible POS labels. Only when the correct label is selected does an agent move to the next state, receiving a reward $1$. All incorrect labels result in a transition to the terminal state with a reward $0$. For a detailed example, please see Appendix.

In this setting, for each state-action pair $(s,a)$, the action-value function $Q(s,a)$ is the expected discounted number of steps that the agent will be able to correctly label the next word, from the current state $s$ if it took labeling action $a$.
If this number for an action $a$ is the largest among all others, it suggests that this action $a$ is the correct label.
Additionally, the magnitudes of $Q(s,a)$ for the possible actions provide information about ambiguity: if all but one actions have $Q(s,a)=0$, then the model is quite sure that the maximal action is the correct one; otherwise, some other actions should be considered. 
Here, the training of a POS tagger is equivalent to finding an optimal policy to maximize the expected return by choosing the correct label for each word. 

Note that here the reward signals are dense (defined for each token). While learning from sparse reward signals are hard~\cite{riedmiller2018learning,agarwal2019learning}, our dense reward setting makes the task easier to solve. 
In our experiments, we therefore simply apply Q-learning~\citep{watkins1989learningfd} to learn an optimal policy, but other RL algorithms could also be applied.Following this pattern, other sequence labeling tasks can also be formulated as RL problems.
In contrast to P predictions (i.e. probabilities), we call the action-value functions \textbf{Q predictions}.  Note that Q predictions include discounted future information, whereas P predictions are probabilities for the next word only.

\section{Predictive Representation Learning}\label{sec:prl}

%In this section, we first introduce label trace (T), which encodes observed sequence labels for future usage. Then we propose our method Predictive Representation Learning (PRL) to solve LM utilizing predictive representations.

\begin{figure}[tb]
\vskip 0.2in
\begin{center}
  \centerline{\includegraphics[width=0.8\columnwidth]{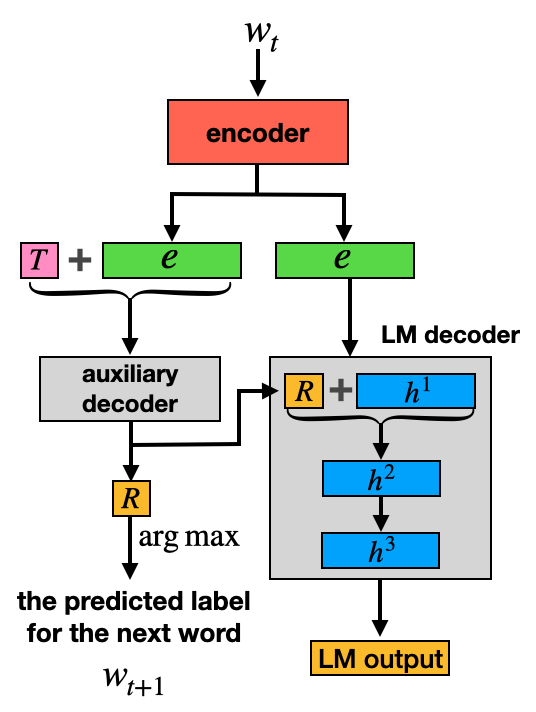}}
  \caption{The model structure of PRL. The encoder (the red rectangle) is shared by both tasks while each task has its own decoder. Each word $w$ is encoded by the shared encoder into a word embedding $e$. The label trace $T$ (the pink square) is concatenated with the word embedding $e$ to form a new word embedding $T+e$. The auxiliary decoder decodes $T+e$ into predictive representations $R$ (the yellow square). For the LM decoder, $R$ is concatenated with the first hidden state $h^{1}$ inside the LM decoder to form a new hidden state $R+h^{1}$. The multi-layer LSTM proceeds as is typical in a language model.}
  \label{fig:model_prl}
\end{center}
\vskip -0.2in
\end{figure}

In this section, we develop a new RNN model with encoder-decoder structure called Predictive Representation Learning (PRL) for the main task of LM. 
% In general, PRL is a framework designed for auxiliary task learning.
% Here, we consider the scenario where LM is the main task.
For the auxiliary task, we consider two sequence labeling tasks -- POS tagging and NER.
Though in principle, other appropriate sequence labeling tasks could be used, we experimented with POS tagging and NER mainly because they are complementary to LM and have smaller prediction spaces.
POS/NER taggers usually predict the POS/NER label of the current word ($w_{t}$), but a language model predicts the identity of the next word.
To align with LM, our POS/NER tagger predicts the POS/NER label for the $\emph{next}$ word ($w_{t+1}$) \emph{without observing it}.
We present our model in Figure~\ref{fig:model_prl}. It consists mainly of three parts - the shared encoder, the auxiliary decoder, and the LM decoder.

Given an input sentence, we feed it as a sequence of words into the \textbf{shared encoder} (the red rectangle in Figure~\ref{fig:model_prl}) where each word $w$ is encoded to a word embedding $e$ (Figure~\ref{fig:model_prl}, green rectangle). The encoder is an embedding layer (i.e. a lookup table that stores word embeddings of a vocabulary set).

The \textbf{auxiliary decoder} is usually an RNN. The input consists of the word embedding and what we call a \textbf{label trace} (described in more detail at the end of this section). The label trace explicitly encodes historical labels (e.g. the POS tags of past words).
We concatenate the label trace vector $T$ (Figure~\ref{fig:model_prl}, pink square) with the word embedding ($e$) to form a new word embedding (i.e. $T+e$) which contains historical label information.  
The auxiliary decoder decodes $T+e$ into predictive representations $R$ (Figure~\ref{fig:model_prl}, yellow square). Depending on the exact form of predictive representations, $R$ can be P predictions or Q predictions. To generate the predicted label, we take $\arg\max$ over $R$.
%Each element in $Q$ is the action-value for taking an action $a \in \Actions$ where $\Actions$ is the POS label set. Thus, the length of this vector is the number of actions in $\Actions$. The action-value for each label represents an estimate of the goodness of that label.

The \textbf{LM decoder} is a 3-layer LSTM.
After the auxiliary decoder computes $R$, it is then concatenated with the first hidden state $h^{1}$ inside the LSTM to form a new hidden state $R+h^{1}$. 
This new hidden state, along with its predictive component, is the input to the next LSTM layer. 
The rest of the computation proceeds as is typical in a multi-layer LSTM and produces a prediction for the next word.
Following previous work, cross-entropy is the loss function.

Finally, an important detail for this architecture is the label trace given to the auxiliary encoder. A label trace is the sum of exponentially discounted one-hot vectors. Each element of the trace vector is a summary of the past observations of a particular label type. 
Formally, let $y_1$, $\ldots$, $y_t$ be the label sequence chosen by the optimal policy. The label trace $T_t(y)$ for each label $y$ is the exponentially discounted count of $y$ over the already observed words, defined as
\begin{equation}
T_{t}(y) \doteq \sum_{k=1}^{t} \gamma^{t-k} \, \mathbbm{1}{(y_{k} = y)}
\end{equation}
where $\gamma \in [0,1]$ and $\mathbbm{1}$ is an indicator function. Higher trace values indicate this label was seen more frequently, recently.

In our experiments, we apply label traces to improve the performance of an auxiliary task.\footnote{We tried the cases that the LM decoder or both decoders had access to label traces. However, there was no significant performance improvement. This is within expectation since the label traces encode labels of an auxiliary task, so they are easier to use by the auxiliary decoder.} Only the auxiliary decoder has access to these label traces, but not the LM decoder, as shown in Figure~\ref{fig:model_prl}.  The label traces boost the performance of LM indirectly by improving the quality of predictive representations, as we will show in the following experiments.

\section{Experiments}

For the following experiments, LM is the main task while POS tagging is the auxiliary task. To align with the goal of LM, we change the goal of POS tagging to predicting the correct POS label for the $\emph{next}$ word instead of the $\emph{current}$ word. 
We perform our experiments on two widely used LM benchmarks - Penn Treebank (PTB) \citep{marcus1993building} and WikiText-2 (WT2) \citep{merity2016pointer}.
For PTB, we used the version pre-processed by \citet{mikolov2011empirical}. This dataset contains 929k training words, 73k validation words, and 82k test words.
It has a vocabulary of 10,000 words, and words not in the vocabulary are replaced by $\langle unk \rangle$ tokens. 
We extracted the corresponding POS labels from the original PTB dataset.
Since the original PTB dataset has no NER labels, we used the English NER tagger in spaCy to label sentences in PTB.~\footnote{https://github.com/explosion/spaCy}

WT2 is created from Wikipedia articles with over 30,000 words. 
It is about twice the size of the PTB dataset and contains 2088k training words, 217k validation words, and 245k test words.
WT2 has no NER or POS tags, so again we generate tags with the English POS and NER taggers in spaCy.

\begin{figure}[tb]
\vskip 0.2in
\begin{center}
  \centerline{\includegraphics[width=\columnwidth]{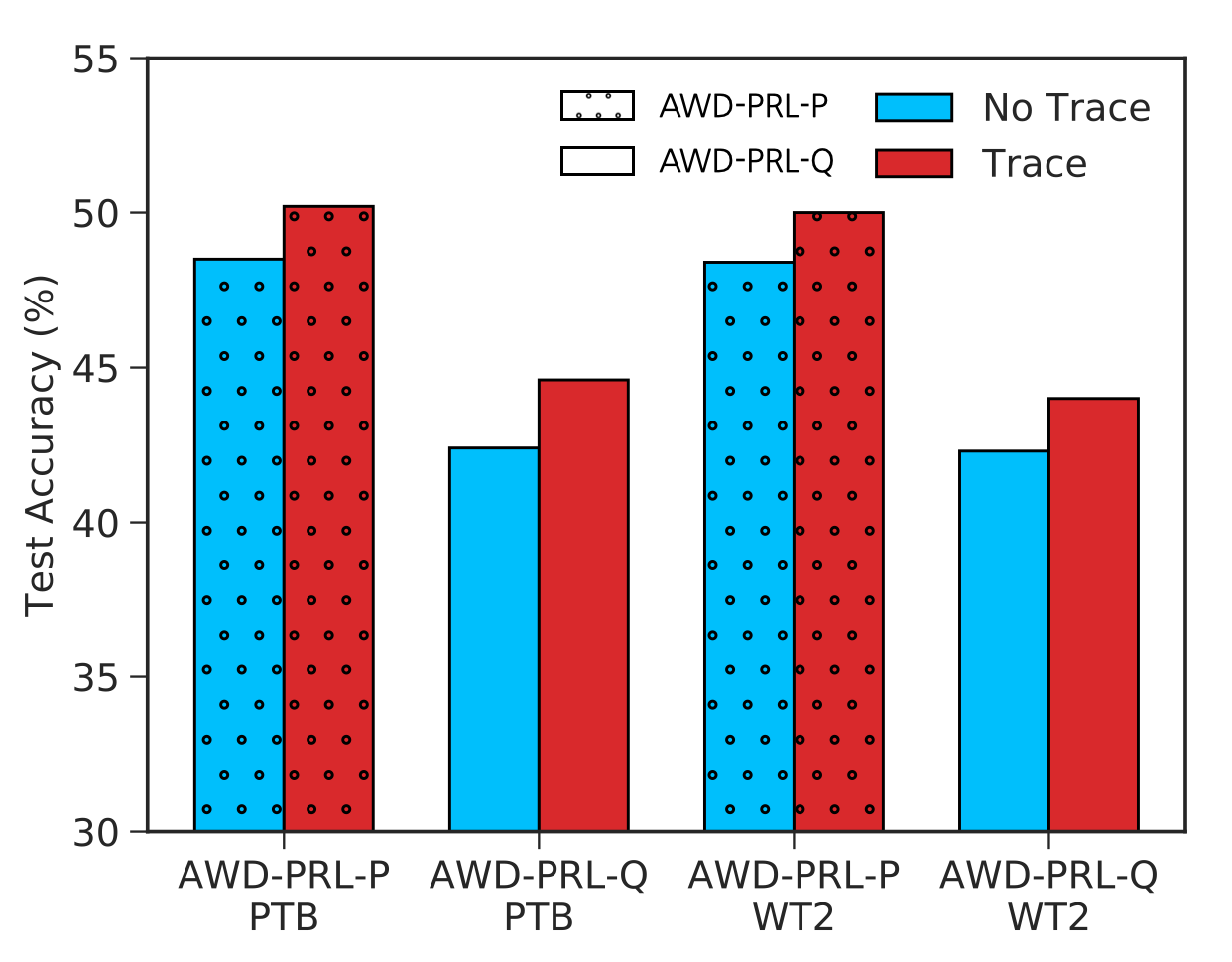}}
  \caption{The comparison of POS tagging accuracy on PTB and WT2 using AWD-PRL-P and AWD-PRL-Q, with and without the label trace. Higher is better. All results were averaged over 5 run. The standard errors are in the range of 0.01\%-0.06\%, and so are to small to be visible on the plot. Overall, the experiments show that the label trace component significantly improves POS prediction.}
  \label{fig:pos}
\end{center}
\vskip -0.2in
\end{figure}

\begin{figure}[tb]
\vskip 0.2in
\begin{center}
  \centerline{\includegraphics[width=\columnwidth]{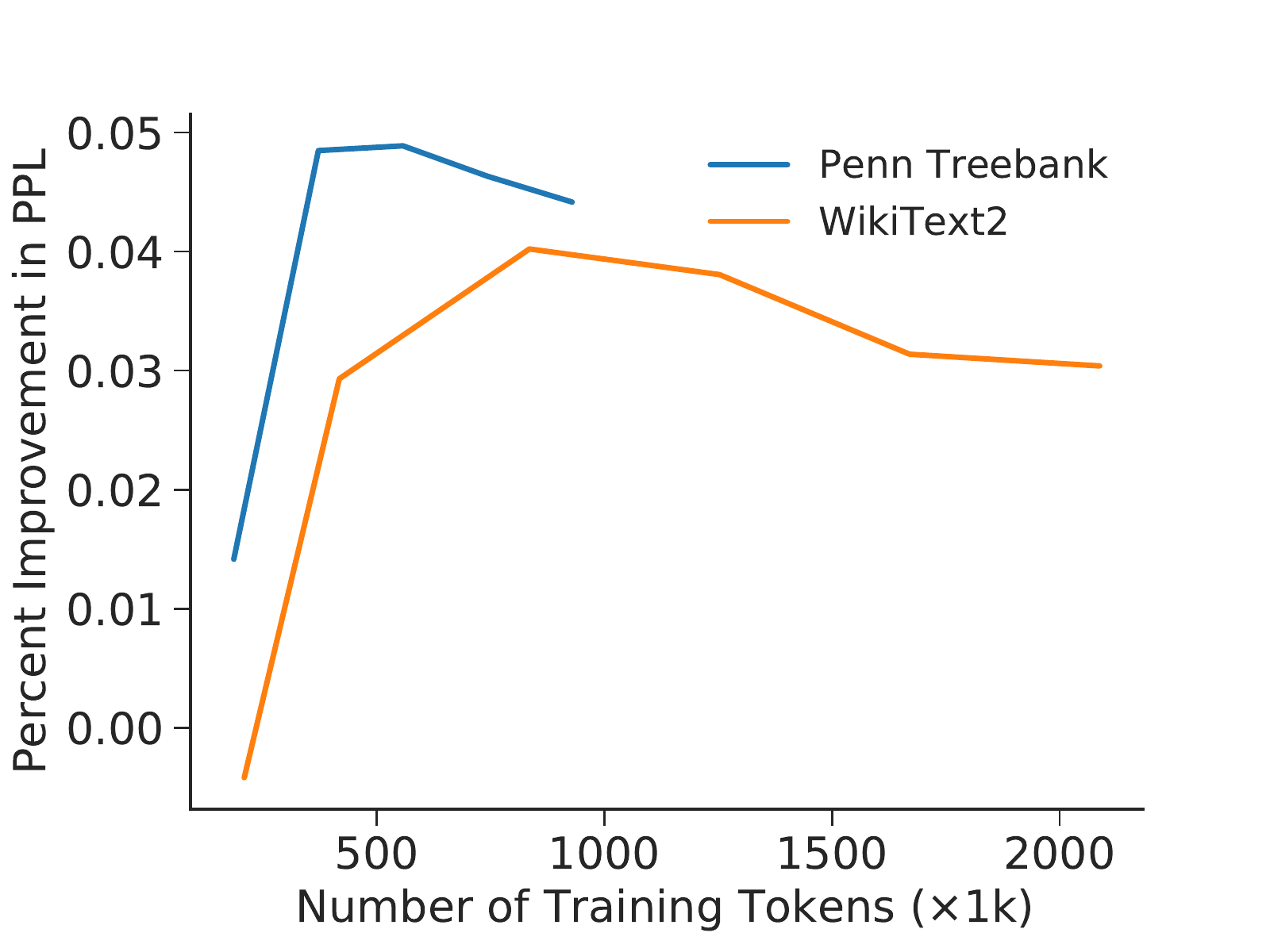}}
  \caption{The percent change in PPL of AWD-PRL-Q over AWD on the test set of PTB and WT2, as a function of dataset size (measured by the number of training tokens). Average over 5 runs is reported. PRL improves language models trained with datasets at different scales.}
  \label{fig:ppl_percent}
\end{center}
\vskip -0.2in
\end{figure}

\subsection{PRL's Effect on LM}
\label{sec:lm}

In this section, we show that the performance of LM can be further improved when predictive representations are incorporated.
% For the following experiments, LM is the main task while POS tagging and NER are two auxiliary tasks.

We first integrate ASGD Weight-Dropped LSTM (AWD-LSTM) \citep{merity2018regularizing} into PRL. We use AWD as a shorter name of AWD-LSTM.
% AWD is an LSTM based language model improved with many advanced techniques, such as DropConnect, NT-ASGD, randomized-length backpropagation through time, embedding dropout, activation regularization, and temporal activation regularization \citep{merity2018regularizing}.
Our models that combine AWD with PRL are called AWD-PRL-P and AWD-PRL-Q; AWD-PRL-P uses P predictions (i.e. probabilities) as predictive representations while AWD-PRL-Q uses Q predictions (i.e. action-value functions). %We will compare the effect of two forms of predictive representations on LM performance.

For the LM decoder, we used the same hyperparameters and optimization settings provided in the official codebase.~\footnote{https://github.com/salesforce/awd-lstm-lm}
The auxiliary decoder is a 2-layer LSTM with a hidden size of 380 and a linear layer neural network.

%For AWD-PRL-Q, the parameters in the POS decoder were optimized by the square loss $L = \frac{1}{|\Actions|} \sum_{a \in \Actions}(Y_t - Q(s,a;\theta))^2$. The parameters in AWD-PRL-P were optimized by cross-entropy loss.
During training, we employed an alternating training approach. For a selected task, a batch of data samples of this task is used to update the parameters in the corresponding decoder and the shared encoder by gradient descent. Next, we switched to a different task and continued this training process until convergence. 
The best learning rate was selected from \{30, 28, 26, 24, 22, 20, 18, 16, 14, 12, 10, 8, 6\} based on validation performance.
The discount factor $\gamma$ was chosen from \{0, 0.5, 0.67, 0.8, 0.9, 0.99\}.
We trained all models for 5 different random initializations.
Following \citet{yang2018breaking}, we removed finetuning to reduce training time. All models were trained for 500 epochs on PTB and 750 epochs on WT2.
Models were evaluated using perplexity (PPL, lower is better).
We reproduced the original results with AWD's official codebase.
More implementation details (runtime, model size etc.) can be found in Appendix.

Table~\ref{tb:ptb_awd} \& \ref{tb:wt2_awd} show the averaged results of all models over 5 runs on PTB and WT2, respectively.
Note that our reproduced results are slightly worse than the results reported in \citet{merity2018regularizing} since we report the averaged perplexities over 5 runs without further finetuning the models.
We report results with and without applying a neural cache. Neural caches keep past hidden activations for later use which greatly boost prediction performance \citep{grave2016improving}.
On both datasets, AWD-PRL-P and AWD-PRL-Q outperform AWD, with and without applying a neural cache.
This supports our hypothesis that constraining the learned representation to be simple predictions improves model performance.
It also confirms that predicted POS/NER tags improve the performance of LM, whether the information itself is represented by P predictions or Q predictions. Compared to NER, using POS tagging as the auxiliary task results in a greater reduction of PPL.
%Furthermore, the perplexities of AWD-PRL-Q are consistently lower than the perplexities of AWD-PRL-P. This is evidence that the Q predictions' representation of future information (as a form of predictive representations) is more beneficial to LM than simple probabilities (P predictions). 

\begin{table*}[tb]
\caption{The comparison of test PPLs on PTB for models based on AWD. For test PPL, lower is better. The $\Delta$ columns show the improvements of AWD-PRL-P/Q in terms of PPL, compared with AWD. We report results with and without a neural cache. All results were averaged over 5 runs with the standard errors reported. For both auxiliary tasks, PRL improves the performance of AWD, with and without a neural cache.}
\label{tb:ptb_awd}
\vskip 0.15in
\centering
\begin{tabular}{lccccc}
\toprule
\multirow{2}{*}{Auxiliary Task} & \multirow{2}{*}{Model} & \multicolumn{2}{c}{No Neural Cache} & \multicolumn{2}{c}{Neural Cache} \\
~ & ~ & Test PPL & $\Delta$ & Test PPL & $\Delta$ \\
\midrule
/   & AWD       & 58.48 $\pm$ 0.06 &  /   & 54.36 $\pm$ 0.07 & /    \\
NER & AWD-PRL-P & 56.84 $\pm$ 0.03 & 1.64 & 52.74 $\pm$ 0.03 & 1.62 \\
NER & AWD-PRL-Q & 57.06 $\pm$ 0.05 & 1.42 & 52.97 $\pm$ 0.05 & 1.39 \\
POS tagging & AWD-PRL-P & 56.31 $\pm$ 0.06 & 2.17 & 52.01 $\pm$ 0.05 & 2.35 \\
POS tagging & AWD-PRL-Q & \textbf{55.90 $\pm$ 0.06} & \textbf{2.58} & \textbf{51.90 $\pm$ 0.06} & \textbf{2.46} \\
\bottomrule
\end{tabular}
\vskip -0.1in
\end{table*}

\begin{table*}[tb]
\caption{The comparison of test PPLs on WT2 for models based on AWD. For test PPL, lower is better. The $\Delta$ columns show the improvements of AWD-PRL-P/Q in terms of PPL, compared with AWD. We report results with and without a neural cache. All results were averaged over 5 runs with the standard errors reported. For both auxiliary tasks, PRL improves the performance of AWD, with and without a neural cache.}
\label{tb:wt2_awd}
\vskip 0.15in
\centering
\begin{tabular}{lccccc}
\toprule
\multirow{2}{*}{Auxiliary Task} & \multirow{2}{*}{Model} & \multicolumn{2}{c}{No Neural Cache} & \multicolumn{2}{c}{Neural Cache} \\
~ & ~ & Test PPL & $\Delta$ & Test PPL & $\Delta$ \\
\midrule
/   & AWD       & 68.85 $\pm$ 0.11 & /    & 54.24 $\pm$ 0.07 & /    \\
NER & AWD-PRL-P & 67.14 $\pm$ 0.09 & 1.71 & 52.91 $\pm$ 0.07 & 1.33 \\
NER & AWD-PRL-Q & 67.03 $\pm$ 0.03 & 1.82 & 52.96 $\pm$ 0.01 & 1.28 \\
POS tagging & AWD-PRL-P & 67.52 $\pm$ 0.06 & 1.33 & 52.86 $\pm$ 0.05 & 1.38 \\
POS tagging & AWD-PRL-Q & \textbf{66.75 $\pm$ 0.07} & \textbf{2.10} & \textbf{52.58 $\pm$ 0.05} & \textbf{1.66} \\
\bottomrule
\end{tabular}
\vskip -0.1in
\end{table*}

\begin{figure}[tb]
\vskip 0.2in
\begin{center}
  \includegraphics[width=\columnwidth]{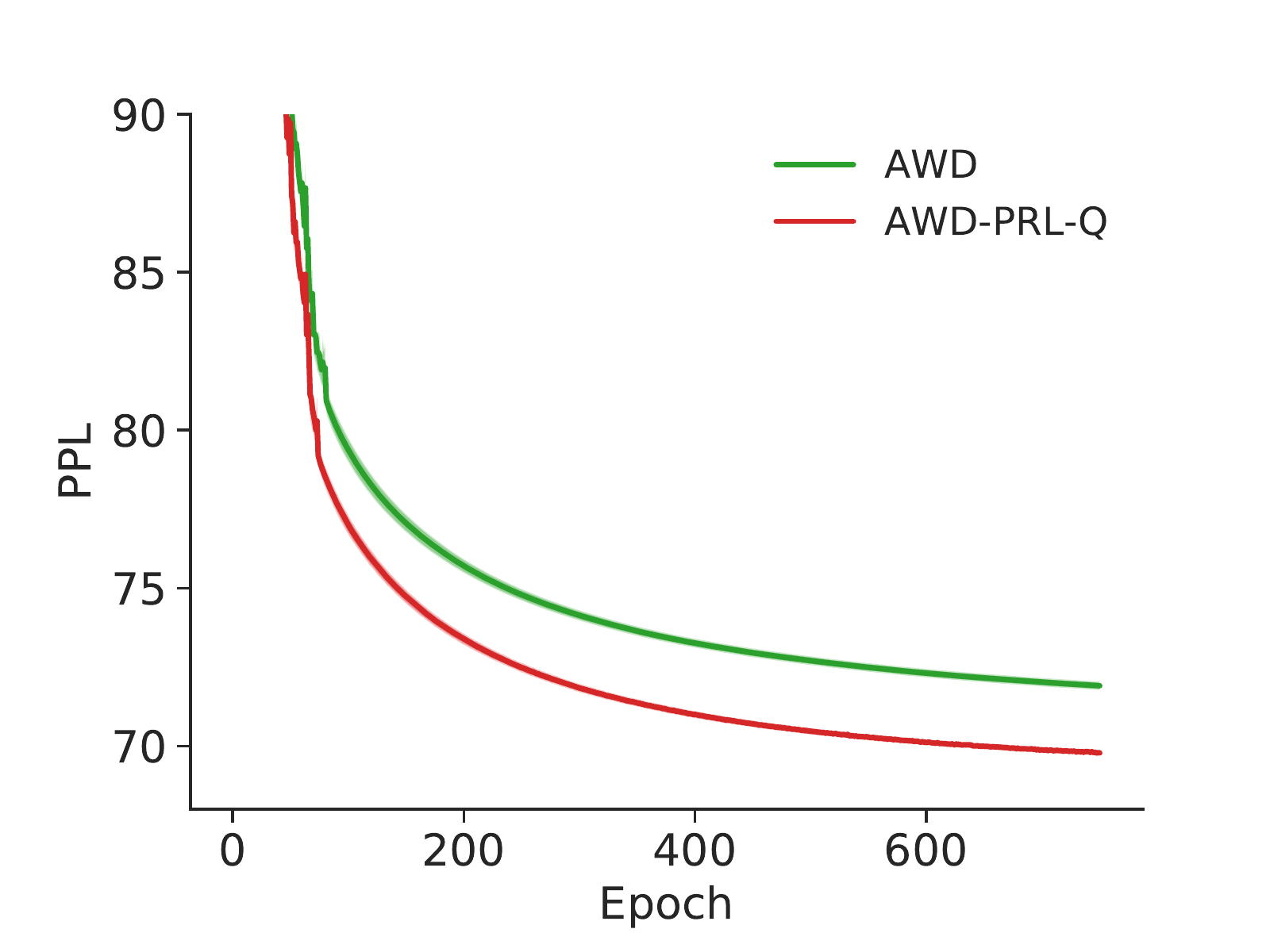}
  \caption{The validation PPL for AWD and AWD-PRL-Q on WT2 during training. Lower is better. Average over 5 runs is shown, shaded regions represent standard errors. AWD-PRL-Q converges faster than AWD.}
  \label{fig:lc_awd_WT2}
\end{center}
\vskip -0.2in
\end{figure}

To test the generalization of PRL, we incorporated it into another language model, Mixture of Softmaxes (AWD-LSTM-MoS) \citep{yang2018breaking}. Here we use MoS as a shorter name for AWD-LSTM-MoS.
% In practice, the capacity of softmax-based language models is limited by the softmax function, which impedes our ability to model highly context-dependent patterns in natural languages. \citet{yang2018breaking} introduce latent variables into an RNN language model and propose the MoS method to improve the expressiveness of the softmax function.
Our models that combine AWD with PRL are called MoS-PRL-P and MoS-PRL-Q, using P predictions and Q predictions, respectively.

We used a similar training process and hyperparameter tuning as in the previous AWD experiment.
Models were trained to convergence for 800 epochs on both datasets. For AWD, POS tagging was clearly a better auxiliary task; thus for these experiments, we only explore POS tagging as an auxiliary task. %The runtime and the size of each model are in Appendix.
We reproduced the original results using the official codebase.~\footnote{https://github.com/zihangdai/mos}

Table~\ref{tb:pos_mos} shows the averaged results of the two models over 5 runs as well as the standard errors.
Results with dynamic evaluation \citep{krause2018dynamic} (column Dyneval) and without dynamic evaluation (column No Dyneval) are reported. Dynamic evaluation helps a model adapt to recent sequences and improves its performance.

MoS-PRL-P and MoS-PRL-Q outperform MoS on PTB and WT2 consistently without dynamic evaluation. With dynamic evaluation, the performance gaps of these three models narrow. Though MoS-PRL-P and MoS-PRL-Q on PTB with dynamic evaluation outperform the baseline MoS, they result in a slight increase in PPL on WT2 (though the PPLs are within a standard deviation of each other).
Overall, the results show that PRL improves the performance of LM.

\begin{table*}[tb]
\caption{The comparison of test PPLs on PTB and WT2 for models based on MoS, using POS tagging as an auxiliary task. For test PPL, lower is better. The $\Delta$ columns show the improvements of MoS-PRL-Q in terms of PPL, compared with MoS. We reported results with dynamic evaluation (Dyneval) and without (No Dyneval). All results were averaged over 5 runs with the standard errors reported. Overall, PRL improves the performance of MoS.}
\label{tb:pos_mos}
\vskip 0.15in
\centering
\begin{tabular}{lccccc}
\toprule
\multirow{2}{*}{Dataset} & \multirow{2}{*}{Model} & \multicolumn{2}{c}{No Dyneval} & \multicolumn{2}{c}{Dyneval} \\
~ & ~ & Test PPL & $\Delta$ & Test PPL & $\Delta$ \\
\midrule
PTB   & MoS     & 56.44 $\pm$ 0.06 &  /   & 49.99 $\pm$ 0.05 & /     \\
PTB & MoS-PRL-P & 54.07 $\pm$ 0.08 & 2.37 & 48.86 $\pm$ 0.07 & 1.13 \\
PTB & MoS-PRL-Q & \textbf{53.88 $\pm$ 0.06} & \textbf{2.56} & \textbf{48.21 $\pm$ 0.04} & \textbf{1.78}  \\
\midrule
WT2   & MoS       & 64.92 $\pm$ 0.18 & /    & \textbf{44.08 $\pm$ 0.07} & /   \\
WT2   & MoS-PRL-P & 64.82 $\pm$ 0.13 & 0.10 & 44.19 $\pm$ 0.04 & -0.11 \\
WT2   & MoS-PRL-Q & \textbf{64.39 $\pm$ 0.14} & \textbf{0.53} & 44.13 $\pm$ 0.08 & -0.05 \\
\bottomrule
\end{tabular}
\vskip -0.1in
\end{table*}

\subsection{Analysis of the Convergence Rate and the Influence of the Dataset Size}

\begin{figure}[tb]
\vskip 0.2in
\begin{center}
  \centerline{\includegraphics[width=\columnwidth]{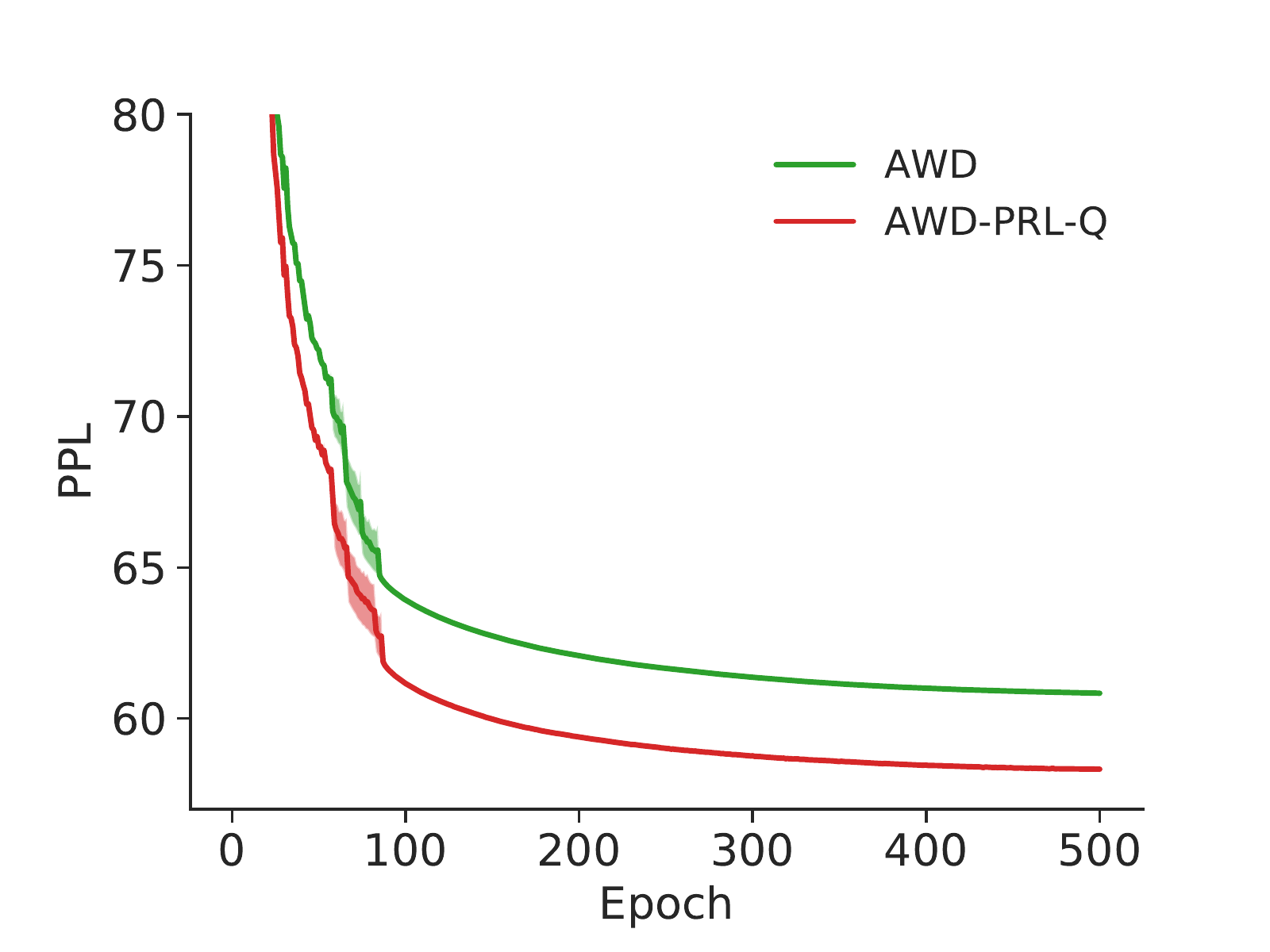}}
  \caption{The validation PPL for AWD and AWD-PRL-Q on PTB during training. Lower is better. Average over 5 runs is shown, shaded regions represent standard errors. AWD-PRL-Q converges faster than AWD.}
  \label{fig:lc_awd_PTB}
\end{center}
\vskip -0.2in
\end{figure}

\begin{figure}[tb]
\vskip 0.2in
\begin{center}
  \centerline{\includegraphics[width=\columnwidth]{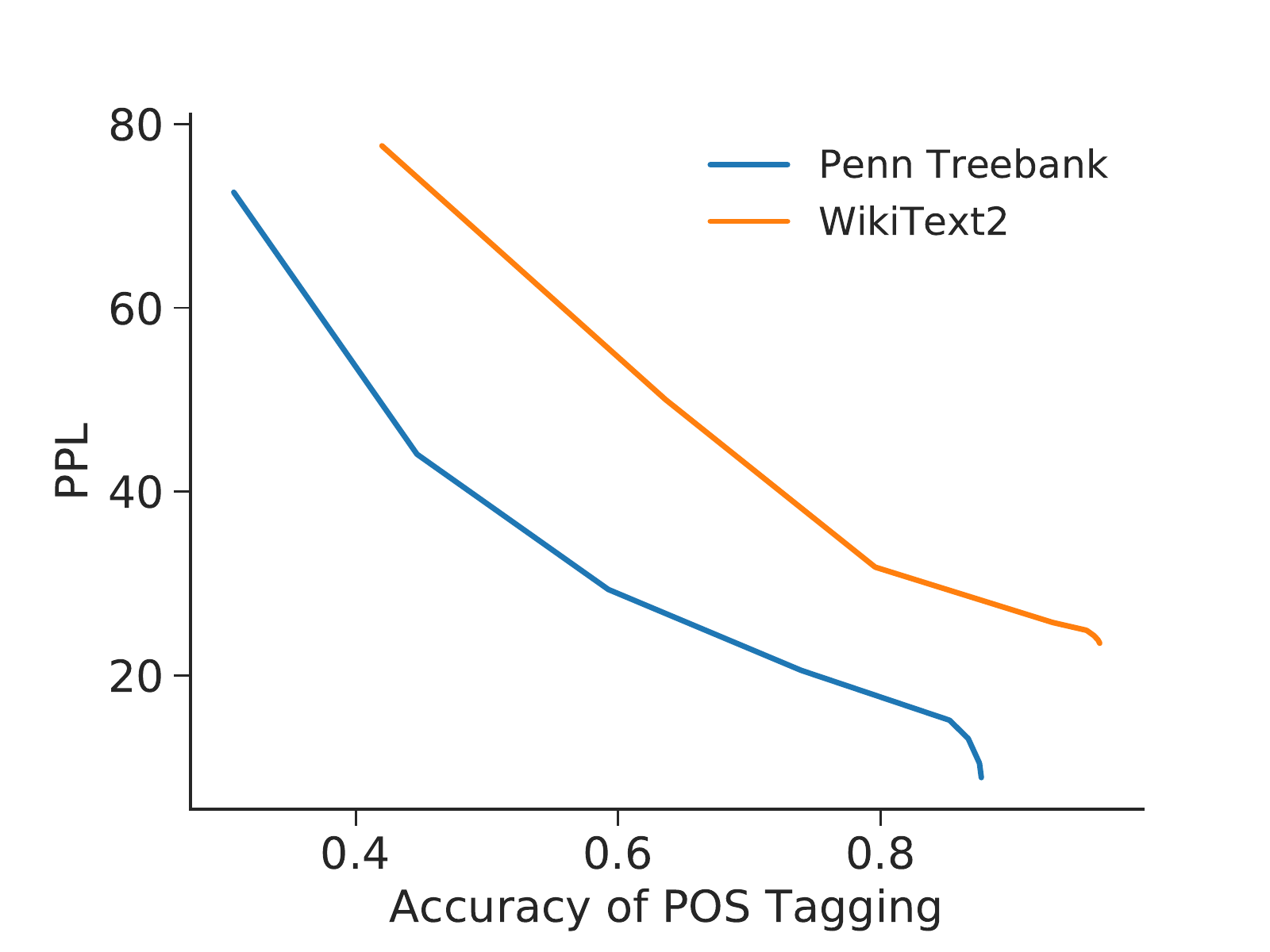}}
  \caption{The relation between the accuracy of POS tagging and PPL of LM on PTB and WT2 for MoS-PRL-Q. The PPL drops almost linearly as the POS tagging accuracy increases.}
  \label{fig:cheating_mos}
\end{center}
\vskip -0.2in
\end{figure}

We have shown that PRL improves the performance of the final trained language models. Because PRL helps to guide representation learning, we hypothesized it would also improve the convergence rate and sample efficiency. 
Figure \ref{fig:lc_awd_PTB} shows the validation PPL while training AWD and AWD-PRL-Q on PTB, with POS tagging as the auxiliary task.
Notice that the PPL of AWD after 500 epochs is achieved by AWD-PRL-Q after only 100 epochs.
The results for WT2 are similar as shown in Figure~\ref{fig:lc_awd_WT2}.

PRL helps language models converge faster, but what is convergence like with less data?
To study the influence of dataset size on LM performance, we trained AWD and AWD-PRL-Q on a subset of the training data, with POS tagging as the auxiliary task. We created several smaller PTB and WT2 datasets and trained models on them. 
In Figure~\ref{fig:ppl_percent}, we see improvements on PTB, even when we include only 250k tokens. The performance of AWD-PRL-Q on WT2 with 250k tokens is slightly worse than AWD, but improvements are seen with just slightly more tokens. The percent improvement in PPL can approach 5\%, and it appears to stabilize at 3\% even as the dataset size surpasses 2M tokens. Thus, PRL can improve language models trained with smaller datasets, and even when the datasets are quite large, there still exists a significant improvement.

\subsection{Performance of POS Tagging}
We were also interested in the performance of POS tagging, and the impact of the label trace on that accuracy.
To study this, we removed the label trace components from AWD-PRL-P and AWD-PRL-Q to create two new models, denoted as \emph{AWD-PRL-P (no T)} and \emph{AWD-PRL-Q (no T)} respectively.
We trained the models as described in Section~\ref{sec:lm} and reported the POS tagging accuracy of each model.

Figure~\ref{fig:pos} shows the test performance of AWD-PRL-P and AWD-PRL-Q on PTB and WT2, with and without the label trace. All results were averaged over 5 runs with the standard errors reported.

Notice that predicting the POS label for the next word is much harder than predicting the POS label for the current word. 
The accuracy of POS tagging for the next word is barely 50\% while state-of-the-art accuracy of English POS tagging for the \emph{current} word is close to 100\%~\citep{bohnet2018morphosyntactic}.

The ablation experiments also show that the label trace component significantly improves POS prediction.
% This suggests that forcing the representation to explicitly encode historical information improves future prediction.
This suggests that label traces given to the auxiliary decoder are believed to be useful.
Although both AWD-PRL-P and AWD-PRL-Q can \emph{learn to} extract and retain historical information, the explicit encoding of history helps. More ablation results for the label trace in AWD-PRL-Q can be found in Appendix.

Note that, in terms of POS tagging, AWD-PRL-P outperforms AWD-PRL-Q on both datasets as shown in Figure~\ref{fig:pos}. In Section \ref{sec:lm}, we showed that AWD-PRL-Q is better than AWD-PRL-P for LM. Thus, AWD performs better for LM when supported by a less accurate POS tagger. As we will see, this differs from the results in Section~\ref{sec:cheating}, which show that (in general) higher POS tagging accuracy leads to better performance of LM.
Recall that Q predictions from AWD-PRL-Q encode not only the prediction of the immediate next POS label, but all labels following the current word. In comparison, the probability distribution in AWD-PRL-P only tells us about the POS label of the immediate \emph{next word}. A plausible explanation is that AWD-PRL-Q's additional \emph{future} predictions provide the performance boost.

\subsection{Effect of POS Tagging Accuracy on PRL}
\label{sec:cheating}

We have shown that the performance of LM is improved by PRL. Note that this is true even though the accuracy of the underlying POS tagging is as low as 44\%.
This inspired us to ask: what would the impact of PRL be if POS tagging was better? How does a language model improve as POS tagging improves?

Current state-of-the-art methods for predicting a POS label given the representation of the current word (i.e. the word to be tagged) are extremely accurate~\citep{bohnet2018morphosyntactic}.
Recall that our model predicts the POS label of the next word without access to that word's embedding. 
We leveraged these two facts in order to study the effect of POS tagging accuracy on MoS-PRL-Q.
We developed an \emph{oracle model}, wherein the POS tagger, trained to predict the POS label of the next word, actually has access to the embedding of that next word (rather than access only to the current word). We are totally aware that this oracle access can leak some information about the next word to the LM decoder through the auxiliary decoder, and it is not a valid way to build a real language model scientifically. But, because the upper limit of these oracle predictions can approach 90\%, we can more fully experiment with the impact of POS tagging accuracy on MoS-PRL-Q and explore the limits of PRL.
To control the accuracy of POS tagging, we varied the size of the hidden layers in the POS decoder (layers with fewer hidden units leads to lower performance). Note that we trained the oracle model with only \emph{10} epochs.

Figure~\ref{fig:cheating_mos} shows how PPL changes as the accuracy of POS tagging changes. 
The PPL decreases almost linearly as the accuracy of POS tagging increases to 80.
This strongly supports our hypothesis that constraining representations to be predictive can improve the performance of LM. Moreover, it hints at a promising future for PRL. As we improve the accuracy of POS tagging or incorporate other sequence labeling tasks with higher accuracy, we can expect that the performance of LM will improve too.

\section{Conclusion and Future Work}

In this paper, we introduced PRL as an addition to RNN models that learns predictive representations. 
PRL takes inspiration from RL to improve learned solutions for sequence labeling tasks. 
We incorporated PRL into two strong language models and showed PRL improved the performance of both.

Our results point to several options for future work. 
To check the generalization of PRL further, we would like to test it on more datasets, more sequence labeling tasks, more forms of predictive representations, and measure the influence of the number of auxiliary tasks. 
Combining PRL with other RL algorithms (e.g. Actor-Critic~\citep{sutton2011reinforcement}) is also an interesting direction. 
The transformer~\citep{dai2019transformer}, which has shown great improvement for LM, is another ripe area for PRL. 
Furthermore, replacing softmax function with hierarchical softmax~\cite{mohammed2018effectiveness} in PRL may further improve the model performance.

We also note that our work requires additional labeled data to learn predictive representations. 
In the future, it may be possible for algorithms to define their own supervision tasks (self-supervision), and incorporate those predictions into the learned representations. 
As mentioned previously, this occurs naturally in RNNs for LM, but encouraging it more explicitly could leverage the advances we presented here without requiring new labeled datasets. 
However, the NLP community's strong tradition of creating benchmarks (and associated labeled data) means there will likely remain plenty to gain by using supervised learning in PRL.

\newpage
\bibliographystyle{icml2021}
\bibliography{reference}

\onecolumn
\newpage
\twocolumn
\appendix

%In this setting, $Q(s,a)$ is the expected discounted number of steps that a policy using $Q(s,a)$ will be able to correctly label the next word, from the current state $s$ if it took labeling action $a$. If this number is large, it suggests that $a$ is the correct label. Additionally, the magnitudes of $Q(s,a)$ for the possible actions provide information about ambiguity: if all the other actions have $Q(s,a)=0$, then the model is quite sure that the maximal action is the right label; otherwise some of the other actions could be considered. The goal is to find an optimal policy that maximizes the expected target, by choosing the correct label for each word. Following this pattern, other sequence labeling tasks can also be formulated as RL problems.

\section{Additional Results for Convergence Rate and Influence of Dataset Size}

\begin{figure}[htbp]
\vskip 0.2in
\begin{center}
  \includegraphics[width=\columnwidth]{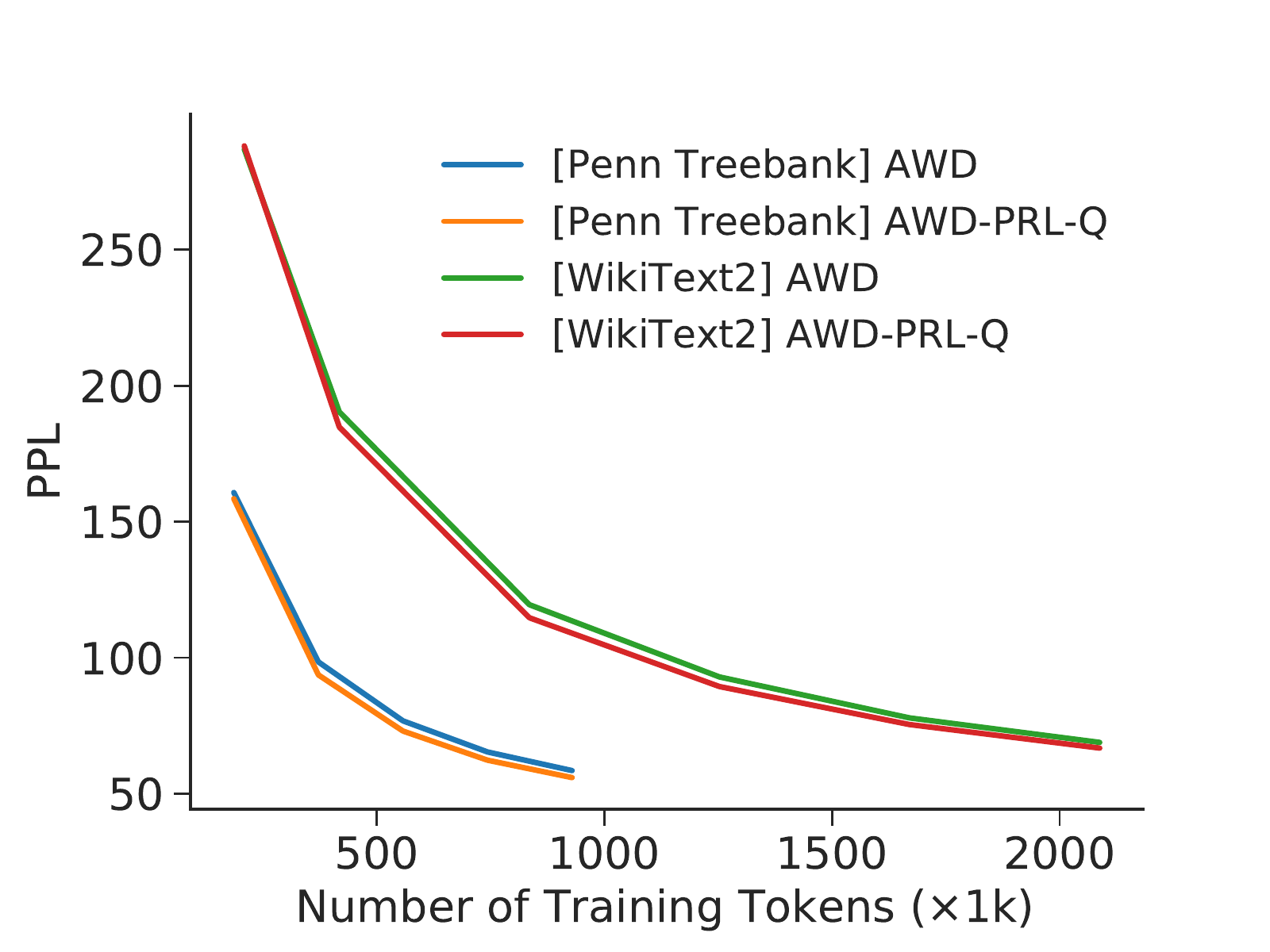}
  \caption{The test PPL as a function of dataset size. Lower is better. Average over 5 runs is reported. PRL improves language models trained with datasets at different scales.}
  \label{fig:ppl_small_data}
\end{center}
\vskip -0.2in
\end{figure}

We report the PPL on the test set as the number of tokens changes for AWD and AWD-PRL-Q on PTB and WT2 (Fig.~\ref{fig:ppl_small_data}).

\section{Model Ablation Analysis}

We have shown that the label trace component improves the POS prediction significantly. We further showed that higher POS tagging accuracy leads to a better language model. Based on these two statements, we draw the corollary that the label trace component should improve the performance of language models. In this section, we conducted an ablation study on both PTB and WT2 to check the corollary. We remove the label trace component from AWD-PRL-Q and the new model is denoted as \emph{AWD-PRL-Q (no T)}.

We report the results in Table~\ref{tb:aux_awd_T}.
The performance of \emph{AWD-PRL-Q (no T)} is significantly worse than AWD-PRL-Q, even worse than AWD.
This supports our corollary that the label trace component can improve the performance of language modeling.

\section{SentEval Results with PRL}

We tested the model representations on the well-known sentence embeddings evaluation benchmark SentEval~\citep{conneau2018senteval}.
It includes binary and multi-class classification, natural language inference, and sentence similarity tasks. 

We compared the representations of our model MoS-PRL-Q and the representations of MoS (both models were trained on WT2).
Table~\ref{tb:senteval_classification} and Table~\ref{tb:senteval_similarity} present the results generated by a logistic regression classifier on SentEval.
We find our model MoS-PRL-Q to have better results compared to MoS which indicates that it generates better representations.

As a comparison, we also include results for Glove~\citep{pennington2014glove} vectors trained on Common Crawl dataset.
Glove outperforms both MoS-PRL-Q and MoS significantly because it was trained on a much larger dataset with 840B tokens while MoS-PRL-Q and MoS were trained on WT2 with only 2M tokens.

\section{Runtime and Model Size}

All models were trained on an NVIDIA Tesla V100-32GB GPU.
We report the training speed and the number of parameters of each model in Table~\ref{tb:speed_size}. All models were trained on an NVIDIA Tesla V100-32GB GPU.

\begin{table}[htbp]
\caption{The training speed and the number of parameters of each model.}
\label{tb:speed_size}
\vskip 0.15in
\centering
\begin{tabular}{llcc}
\toprule
Dataset & Model & Speed & \# Param \\
\midrule
PTB    & AWD           &  38 s/epoch & 35.44 M \\
PTB    & AWD-PRL-P/Q   &  55 s/epoch & 39.20 M \\
\midrule
WT2    & AWD           &  60 s/epoch & 44.76 M \\
WT2    & AWD-PRL-P/Q   &  76 s/epoch & 48.59 M \\
\midrule
PTB    & MoS           &  81  s/epoch & 30.41 M \\
PTB    & MoS-PRL-Q     &  110 s/epoch & 33.95 M \\
\midrule
WT2    & MoS           &  412 s/epoch & 47.17 M \\
WT2    & MoS-PRL-Q     &  476 s/epoch & 50.84 M \\
\bottomrule
\end{tabular}
\vskip -0.1in
\end{table}

\section{Implementation Details}

We release our source code as a part of the supplementary materials. It includes the implementation of AWD-PRL-P/Q and MoS-PRL-P/Q, bounds for each hyperparameter, and the hyperparameter configurations for best-performing models. We apply the grid search for a hyperparameter search. We do not release the PTB since it is licensed. However, WT2 is released.

\begin{table*}[htbp]
\caption{The ablation study of the label trace for AWD-PRL-Q on PTB and WT2. For test PPL, lower is better. \emph{AWD-PRL-Q (no T)} is derived from AWD-PRL-Q by removing the label trace component. The $\Delta$ columns show the improvements of AWD-PRL-Q and \emph{AWD-PRL-Q (no T)} in terms of PPL, compared with AWD. We report results with and without a neural cache (column Neural Cache and column No Neural Cache). All results were averaged over 5 runs with the standard errors reported. Without the label trace, the performance of AWD-PRL-Q is significantly worse.}
\label{tb:aux_awd_T}
\vskip 0.15in
\centering
\begin{tabular}{lccccc}
\toprule
\multirow{2}*{Dataset} & \multirow{2}*{Model} & \multicolumn{2}{c}{No Neural Cache} & \multicolumn{2}{c}{Neural Cache} \\
~ & ~ & Test PPL & $\Delta$ & Test PPL & $\Delta$ \\
\midrule
PTB & AWD              & 58.48 $\pm$ 0.06 &  /   & 54.36 $\pm$ 0.07 & /    \\
PTB & AWD-PRL-Q        & 55.90 $\pm$ 0.06 & 2.58 & 51.90 $\pm$ 0.06 & 2.46 \\
PTB & AWD-PRL-Q (no T) & 58.74 $\pm$ 0.04 & -0.26 & 54.54 $\pm$ 0.03 & -0.18 \\
\midrule
WT2 & AWD              & 68.85 $\pm$ 0.11 & /    & 54.24 $\pm$ 0.07 & /    \\
WT2 & AWD-PRL-Q        & 66.75 $\pm$ 0.07 & 2.10 & 52.58 $\pm$ 0.05 & 1.66 \\
WT2 & AWD-PRL-Q (no T) & 69.29 $\pm$ 0.07 & -0.44 & 54.51 $\pm$ 0.07 & -0.27 \\
\bottomrule
\end{tabular}
\vskip -0.1in
\end{table*}

\begin{table*}[htbp]
\caption{The test accuracies (x100) of the classification tasks in SentEval for Glove, MoS-PRL-Q, and MoS. Higher is better. All results were averaged over 5 runs. For MRPC, both accuracy (the first number) and F1 score (the second number) are reported. MoS-PRL-Q outperforms MoS on all tasks.}
\label{tb:senteval_classification}
\vskip 0.15in
\centering
\begin{tabular}{llllllllll}
\toprule
Model & MR & CR & SUBJ & MPQA & SST-2 & SST-5 &  TREC &  MRPC & SICK-E \\
\midrule
Glove     & 77.1  & 78.78 & 91.07 & 87.59 & 79.68 & 43.8  & 82.8  & 73.78/77.58 & 78.69\\
MoS-PRL-Q & 60.92 & 65.46 & 85.25 & 69.98 & 64.44 & 33.38 & 56.48 & 68.14/79.99 & 69.92\\
MoS       & 54.05 & 63.78 & 72.29 & 68.76 & 58.12 & 29.27 & 39.6  & 66.49/79.87 & 56.97\\
\bottomrule
\end{tabular}
\vskip -0.1in
\end{table*}

\begin{table*}[htbp]
\caption{The evaluation of sentence representations on the semantic textual similarity benchmarks. Higher is better. The numbers reported are Pearson correlations (x100). All results were averaged over 5 runs. Averaged correlation scores are reported for STS'12 to STS'16 which are composed of several subtasks. MoS-PRL-Q outperforms MoS on all tasks.}
\label{tb:senteval_similarity}
\vskip 0.15in
\centering
\begin{tabular}{llllllll}
\toprule
Model  & SST'12 & SST'13 & SST'14 & SST'15 & SST'16 & SICK-R &  SST-B \\
\midrule
Glove     & 52.2  & 49.6  & 54.59 & 56.25 & 51.40 & 79.92 & 64.78 \\
MoS-PRL-Q & 29.32 & 17.45 & 22.56 & 29.71 & 31.41 & 68.41 & 49.66 \\
MoS       & 10.39 & -4.04 & -6.07 & 6.44  & 7.76  & 38.59 & 24.55 \\
\bottomrule
\end{tabular}
\vskip -0.1in
\end{table*}

\end{document}